# The Arabic Noun System Generation[1]

Abdelhadi Soudi, Violetta Cavalli-Sforza and Abderrahim Jamari

In this paper, we show that the multiple-stem approach to nouns with a broken plural pattern allows for greater generalizations to be stated in the morphological system. Such an approach dispenses with truncating/deleting rules and other complex rules that are required to account for the highly allomorphic broken plural system. The generation of inflected sound nouns necessitates a pre-specification of the affixes denoting the sound plural masculine and the sound plural feminine, namely *uwna* and *aAt*, in the lexicon. The first subsection of section one provides an evaluation of some of the previous analyses of the Arabic broken plural. We provide both linguistic and statistical evidence against deriving broken plurals from the singular or the root. In subsection two, we propose a multiple stem approach to the Arabic Noun Plural System within the Lexeme-based Morphology framework. In section two, we look at the noun inflection of Arabic. Section three provides an implementation of the Arabic Noun system in MORPHE. In this context, we show how the generalizations discussed in the linguistic analysis section are captured in Morphe using the equivalencing nodes.

## Introduction

There are three number categories for Arabic nouns (including adjectives): singular (*mufrad*), dual (*mu#anGaY*), and plural (*jam‘*). The plural is further divided into sound (*al-jam‘u Al-sGaAlim-u*), the use of which is practically confined to (at least in the masculine) to participles and nouns indicating profession and habit, and broken (*al-jam‘u l-mukasGaru*) types. Broken plurals are then divided into plurals of paucity (*jam‘u l-qilGa0i*), denoting three to ten items, and plurals of multiplicity (*jam’u l-ka#ra0i*), denoting more than ten items. There are four forms of the plural of paucity and at least 23 forms of the plural of multiplicity (‘abGaAs ^abuw al-su‘uwd 1971). Several singulars have more than one plural form. There are also underived nouns with plural or collective sense (usually indicating a group of animals or plants). These are treated as singular, but may form a ‘singulative’ *(çismu l-waHda0i*) indicating an individual of the group, by attachment of the affix *0*.

## 1. The Arabic Broken Plural System

The Arabic Broken plural system is highly allomorphic: For a given singular pattern, two different plural forms may be equally frequent, and there may be no way to predict which of the two a particular singular will take. For some singulars as many as three further statistically minor patterns are also possible. The range of allomorphy is in general from two to five. For example, a singular noun with the pattern CVCC would have one or two of the following plural patterns: CuCuwC, ^aCCaAC, CiCaAC or ^aCCuC. Examples showing the broken plural of the singular pattern CVCC are as follows:

---

[1] Please cf. the symbols we use in the transliteration at the end of paper.

(1)

| singular | plural | gloss |
|---|---|---|
| wazn | ^awzaAn | "measure" |
| kalb | kilaAb | "dog" |
| 'ayn | 'uyuwn, ^a'yun | "eye" |

**1.1. Some previous accounts of The Arabic Broken Plural System**

The Arabic grammarians'approach to the problem of variation is to list the possible patterns (*^abniya0* or *^awzaAn*} of the broken plural and then to try to determine which singular patterns are most usually associated with each. For each plural pattern, they list a series of possible singular forms (sources or literally 'causes' (*'ilal*) and conditions (š*uruwT*) which determine or limit the association of plural forms with particular singular forms. The conditions are of two types, phonological and semantic. However, the phonological conditions accepted by the grammarians relate only to the phonology of the consonantal root- whether a root has a weak consonant (i.e., glottal stop or glide}, for example. Other factors, such as the quality of the singular stem vowel and whether a form is basic or derived, are not explicitly included in the traditional analysis.

The semantic conditions include such criteria as the referent of the noun is animate or inanimate. The introduction of semantic conditions into the analysis is an assertion that certain variations are distinctive; that is, instead of a single functional category plural marked by several different forms, there are actually several different functional categories, such as 'the plural of animate nouns' vs. 'the plural of inanimate nouns'; 'plural of multiplicity' vs. 'plural of paucity'.

To evaluate the statistical productivity in the Arabic plural system, we use Levy's (1971) study which includes statistical information on common plural types, based on Wehr's (1960) dictionary and Murtonen's (1964) survey based on the dictionary of Lane (1893). Tables 1 and 2 show a general index of the productivity of the major Arabic broken plural forms. The left-most column indicates the singular patterns and the top row indicates the most frequent plural forms. [2] There are two numbers at every co-ordinate where the singular line crosses the plural one. The first indicates the percentage of the particular singular in relation to all singulars of a given plural. The second indicates the percentage of the particular plural as a proportion of all the plurals taken by a given singular type. By way of example, the intersection of the singular pattern CaCC and the broken plural pattern *CuCuwC* in table (1a) shows the numbers 73/49. These numbers indicate that 73 percent of all the singulars of a plural pattern *CuCuwC* have a singular of the pattern *CaCC* and that 49 percent of all singulars with the pattern CaCC will have CuCuwC as their plural pattern.

[2] Levy's data include the sound feminine plural forms which are not represented in Murtonen's data . Certain minor singular types (CiCCaAn, CuCCaAn) are represented in Murtonen's data but not in Levy's. Because Murtonen takes the plural form as a starting point for analysis, the total percentages of plural types for a given singular often fall well short of 100%. Therefore, a fairly complete listing of singulars for each plural is provided., but no specific listing of plurals for each singular is given. Note also that singulars with four consonants are not included in the two figures.

# Table 1: singular/plural distribution based on Levy (1971)

a.

| Pl.<br>Sg. —— | ^aCCuC | CuCuwC | ^aCCaAC | CiCaAC | CiCaC | CuCaC | -aAt<br>(sfp) | CawaACiC |
|---|---|---|---|---|---|---|---|---|
| CaCC | 82 /6 | 73/49 | 26/27 | 29/12 | | | | |
| CiCC | 12/3 | 13/23 | 23/67 | 5/7 | | | | |
| CuCC | 6/2 | 6/16 | 17/73 | 5/9 | | | | |
| CvCvC | | 6/9 | 33/85 | 5/6 | | | | |
| CaCCat | | | | 18/18 | 14/5 | 6/3 | 43/74 | |
| CiCCat | | | | 1/4 | 86/84 | | 2/12 | |
| CuCCat | | | | 6/8 | 6/15 | 94/77 | 6/8 | |
| CaCvCat | | | | 5/18 | | | 13/82 | |
| CaACiCat | | 3/4 | 1/2 | 3/2 | | | 5/16 | 58/84 |
| CaACiC | | | | | | | 2/3 | 42/24 |
| CuCuwCat | | | | | | | 1/86 | |
| CaCaACat | | | | | | | 7/74 | |
| CuCaACat | | | | | | | 4/87 | |
| CiCaACat | | | | | | | 4/44 | |
| CaCuwCat | | | | | | | 0/14 | |
| CaCiyCat | | | | | | | | |
| CuCuwC | | | | | | | 1/100 | |
| CaCaAC | | | | | | | 4/27 | |
| CuCaAC | | | | | | | 1/16 | |
| CiCaAC | | | | | | | 6/20 | |
| CaCuwC | | | | 21/16 | | | | |
| CaCiyC | | | | | | | | |

a'.

| Pl.<br>Sg. —— | CuCCaAC | CuCCaC | CaCaCat | CuCa©at | CaCaA@iC | aCCiCat | CuCuC |
|---|---|---|---|---|---|---|---|
| CaCC | | | | | | | |
| CiCC | | | | | | | |
| CuCC | | | | | | | |
| CvCvC | | | | | | | |
| CaCCat | | | | | | | |
| CiCCat | | | | | | | |
| CuCCat | | | | | | | |
| CaCvCat | | | | | | | |
| CaACiCat | | | | | | | |
| CaACiC | 100/26 | 100/10 | 98/14 | 100/11 | | | 5/2 |
| CuCuwCat | | | | | 0/14 | | |
| CaCaACat | | | | | 4/26 | | |
| CuCaACat | | | | | 1/13 | | |
| CiCaACat | | | | | 11/56 | | |
| CaCuwCat | | | | | 2/86 | | |
| CaCiyCat | | | | | 70/95 | | 7/5 |
| CuCuwC | | | | | | | |
| CaCaAC | | | | | 1/5 | 22/52 | 6/13 |
| CuCaAC | | | | | | 8/48 | |
| CiCaAC | | | | | 1/1 | 48/45 | 38/34 |
| CaCuwC | | | | | 3/20 | 3/11 | 16/63 |
| CaCiyC | | | | | 6/4 | 18/17 | 29/11 |

# Table 2: singular/plural distribution based on Murtonen (1964)

a.

| Pl. / Sg. | ^aCCuC | CuCuwC | ^aCCaaC | CiCaAC | CiCaCat | CiCCaAn | CuCCaAn | CuCC | CiCaC | CuCaC |
|---|---|---|---|---|---|---|---|---|---|---|
| CaCC | 57/9 | 56/31 | 23/23 | 24/13 | 19/2 | 25/3 | 15/2 | 6/2 | | |
| CiCC | 11/4 | 17/21 | 21/46 | 6/8 | 29/6 | 12/3 | | | | |
| CuCC | 8/4 | 7/14 | 13/48 | 5/10 | 29/10 | 7/3 | | | | |
| CaCaC | 8/3 | 5/7 | 20/59 | 6/10 | | 6/2 | 13/5 | | | |
| CaCiC | | | 3/31 | 2/17 | | | | | | |
| CiCaC | | | 1/61 | | | | | | | |
| CuCuC | | | 3/68 | | | | | | | |
| CuCaC | | | | | | 6/19 | | | | |
| ^aCCaC | | | | | | | 15/9 | 44/99 | | |
| CaCCat | | 3/6 | | 13/43 | | | | 2/10 | 17/15 | 5/6 |
| CiCCat | | | 1/5 | 4/13 | | | | | 70/69 | |
| CuCCat | | | 1/4 | 5/12 | | | | 4/5 | | 78/71 |
| CaCaCat | | | 1/16 | 3/32 | | | | | | |
| CaACiCat | | | | | | | | | | |
| CaACiC | | 5/4 | 3/4 | 4/3 | | | 12/3 | 7/4 | | |
| CaCaACat | | | | | | | | | | |
| CiCaACat | | | | | | | | | | |
| CaCuwCat | | | | | | | | | | |
| CaCiyCat | | | | | | | | | | |
| CaCaAC | | | 1/16 | | | | | | | |
| CuCaAC | | | | | | 8/19 | | | | |
| CiCaAC | | | | | | | | 7/10 | | |
| CaCuwC | | | | | | | | 7/14 | | |
| CaCiyC | | 2/1 | 4/6 | 11/10 | | 9/2 | 21/5 | 8/5 | | |

a’

| Pl. / Sg. | CawaACiC | CuCCaC | CuCCaAC | CaCaCat | CuCa©at | CaCaa@iC | ^aCCiCat | CuCuC |
|---|---|---|---|---|---|---|---|---|
| CaCC | | | | | | | 6/1 | 1/2 |
| CiCC | | | | | | | | |
| CuCC | | | | | | | | |
| CaCaC | | | | | | | | 0/2 |
| CaCiC | | | | | | | | |
| CiCaC | | | | | | | | |
| CuCuC | | | | | | | | |
| CuCaC | | | | | | | | |
| ^aCCaC | | | | | | | | |
| CaCCat | | | | | | | | |
| CiCCat | | | | | | | | |
| CuCCat | | | | | | | | |
| CaCaCat | | | | | | | | |
| CaACiCat | 56/90 | 7/4 | | | | | | |
| CaACiC | 35/22 | 77/16 | 86/14 | 79/12 | 89/4 | | | 7/5 |
| CaCaACat | | | | | | 3/67 | | |
| CiCaACat | | | | | | 8/83 | | |
| CaCuwCat | | | | | | 2/63 | | |
| CaCiyCat | | | | | | 51/67 | | 5/7 |
| CaCaAC | | | | | | | 13/31 | 4/19 |
| CuCaAC | | | | | | | 8/34 | |
| CiCaAC | | | | | | 4/6 | 31/32 | 19/34 |
| CaCuwC | | | | | | 4/10 | | 20/46 |
| CaCiyC | | | | 7/1 | | 14/8 | 24/9 | 24/16 |

The basic fact that emerges from these statistical tables is that while the association between singular and plural forms is not random, there is no way to predict exactly which plural pattern a singular will take. The association of certain singular patterns with a given group of plural patterns is largely determined by the morpho-phonological form of the singular. For example, singulars having the form CVVCVC(at) (with a long vowel after the first consonant) are associated with one set of plural forms. Singulars of the form CVC(V)C(at) (without a long vowel) are usually associated with a different group of plurals. Singulars having the form CVCVVC(at) (with a long vowel after the second consonant) take another different set of plurals. The morpho-phonological form of the singular noun, however, does not determine the plural pattern the noun will take.

### 1.2. An LBM Treatment of the Arabic Noun Plural System

The problem of variation exhibited in the Arabic broken plural system can be handled in two ways. We can either provide the broken plural pattern in the lexicon and then a series of morphological rules operate on the singular noun to generate the plural noun in the morphological component. These rules would act at the internal level to convert the singular stem to the plural stem and at the external level to add the inflectional affixes (e.g. Case affixes: nominative, genitive or accusative suffixes). The other approach would be to provide the singular and plural stems in the lexicon and then have inflectional morphology act on these stems (Soudi et al 2001). The first approach would obviously involve several rules, for nouns with a broken plural pattern have in general complex stem alternants as we have shown with the previous analyses.

The multiple-stem approach sounds more promising for our purposes. Nouns with a broken pattern commonly display two major stem alternants: the singular/dual allostem and the plural allostem. To capture the fact that there are two forms and that these forms are systematically distributed, the lexeme is given an inventory of two stems. To ensure that the right stem is used for every morphosyntactic word, the stems are labelled with the number feature such that stems with the singular feature are the ones referred to in inflectional rules acting on the singular stem and those with the plural feature in inflectional rules acting on the plural stem.

In our lexeme-based model, the lexeme is made up of two levels of representation. It has a syntactic level, where the syntactic category “noun” is specified, and a phonological level. The latter is given an inventory of stems on which morphological rules act. The lexeme RAJUL "man", for example, would be represented as in (2):

(2)

RAJUL { syntax: noun
phonology (stem inventory):
stem sg or dual /*rajul*/; stem pl /*rijaAl*/ }

The lexeme RAJUL has the stem inventory: *rajul* (singular/dual stem)& *rijaAl* (broken plural stem).

## 2. Noun and Inflection

In this section, we look at the inflection of the Arabic noun system. Consider some syncretism cases in the noun inflection of Arabic.

In table 3 below, the accusative and genitive Cases are realised homonymously in the sound plural. Table 4 shows that the accusative-genitive homonymy does not take place with broken plurals.

| | | Sound Plural Masculine | | Sound Plural Feminine | |
|---|---|---|---|---|---|
| Definiteness | Case | Singular | Plural | Singular | Plural |
| Indefinite | Nominative | mu‘alGimM | mu‘alGim-uwna | HayawaAnM | HayawaAn-aAtu |
| | Accusative | mu‘alGimF | mu‘alGim-iyna | HayawaAnF | HayawaAn-aAtF |
| | Genitive | mu‘alGimK | mu‘alGim-iyna | HayawaAnK | HayawaAn-aAtF |
| Definite | Nominative | mu‘alGimu | mu‘alGim-uwna | HayawaAnu | HayawaAn-aAtu |
| | Accusative | mu‘alGima | mu‘alGim-iyna | HayawaAna | HayawaAn-aAti |
| | Genitive | mu‘alGimi | mu‘alGim-iyna | HayawaAni | HayawaAn-aAti |

**Table 3: Paradigm of word forms of the sound nouns *mu‘alGim* “instructor” and *HayawaAn* “animal”**[3]

[3] The definiteness marker is not included in the table.

| Definiteness | Case | Singular | Plural |
|---|---|---|---|
| Indefinite | Nominative | rajulM | rijaAlM |
| | Accusative | rajulF | rijaAlF |
| | Genitive | rajulK | rijaAlK |
| Definite | Nominative | rajulu | rijaA1u |
| | Accusative | rajula | rijaAla |
| | Genitive | rajuli | rijaAli |

**Table 4: Paradigm of word forms of the broken noun *RAJUL* "man"**

In the relevant literature, the main morphological distinction in declension is that between the broken plurals and the rest. The data in Table 5.4, however, shows that there is another fundamental distinction between two types of sound plurals. This distinction relates to the spell-out of the number marker and the Case marker in the sound plural masculine and the sound plural feminine. The sound plural masculine and the sound plural feminine have different exponents of Plural. The former has *–uwna* (as in *mu`alGim-uwna*) while the latter has *–aAt-* (as in *Hayawan-aAtu*). The data also show that Case is realized in the two types of sound plural in different positions (*-**i**yna*, uwna, *aAt**i***, *aAt**u***) These facts suggest that the stem, the plural inflection and the Case realization be expressed as shown in (3):

(3)

**<u>Rules for Arabic Noun System Generation</u>:**

**(i) Broken Plural versus Sound Plural**

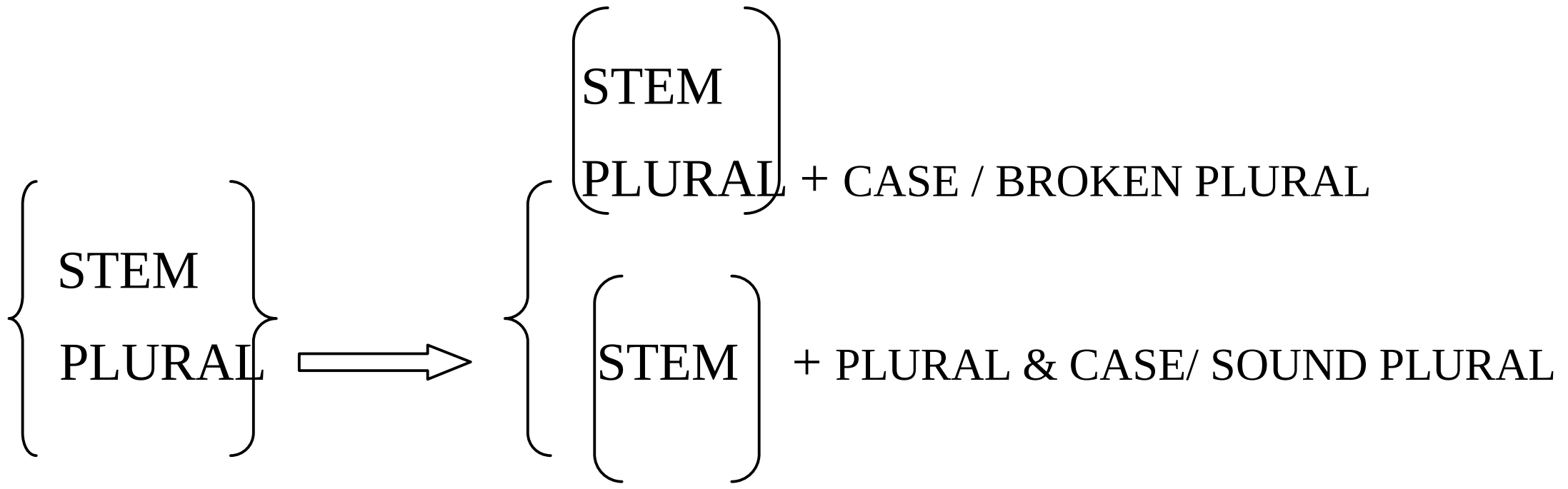

## (ii) Case Realization in the Sound Plural in –*uwna* and Sound Plural in –*aAt*[4]

- **Case**+na $\Longrightarrow$ -uwna Class
- aAt+**Case** $\Longrightarrow$ -aAt Class

## (iii) Genitive-Accusative Take-over for Sound Plurals

$$\text{Noun} + \begin{pmatrix} \text{Plural} \\ \text{Accusative} \end{pmatrix} \Longrightarrow \text{Noun} + \begin{pmatrix} \text{Plural} \\ \text{Genitive} \end{pmatrix}$$

By way of illustration, Table 5 below shows how the above rules apply in the derivation of the inflectional forms for the accusative plural of the two types of the sound plural and the broken plural.

[4] We are not claiming here that Case and Number are realized by different exponents/markers. That is, "Case + na" in (ii) does not mean that *na* is a separate affix (i.e., a number marker) and Case is spelled out with another affix because there are instances where *na* does not surface. A case in point is possessive 7constructions (e.g. *mu'alGimuw al-walad-i* "The boy's instructors").

| | Boken Plural | Sound Plural in -uwna | Sound Plural in -aAt |
|---|---|---|---|
| Morphosyntactic Representation | { rajul<br>PLURAL<br>ACCUSATIVE } | { mu‘alGim<br>PLURAL<br>ACCUSATIVE } | { HaywaAn<br>PLURAL<br>ACCUSATIVE } |
| Broken Plural versus Sound Plural | { RAJUL<br>Plural } + Acc | m. + { Plural<br>Acc } | H. + { Plural<br>Acc } |
| Genitive-Accusative take-over | does not apply | m. + { Plural<br>Gen } | H. + { Plural<br>Gen } |
| Case Realization in the Sound Plural | does not apply | m. + Gen + na | H. + aAt +Gen |
| Spell-out | rijaAl-K | mu‘alGimiyna | HayawanaAti |

**Table 5: The inflectional realization for the accusative plural of the two types of the sound plural and the broken plural**

## 3. The Implementation of the Arabic Noun System in MORPHE

MORPHE (Leavitt 1994) is a tool that compiles morphological transformation rules into into either a word parsing program or word generation program.[5] In this paper, we focus on the use of MORPHE in generation.[6]

---

[5] MORPHE is written in Common Lisp and the compiled transformation rules are themselves a set of Common Lisp functions.

[6] While MORPHE's computational engine is a general one, the morphological rules must be developed for each language. At present, MORPHE is fully functional as a morphological generator, but not as a morphological analyzer. The enhanced version of the MORPHE system (Cavalli-Sforza, 2001) adds allomorph rules and node equivalencing to the original MORPHE system (Leavitt, 1994).

**Inputs and Outputs:**

The input to MORPHE is a feature structure (FS) that describes the item to be generated. An FS is a recursive list structure for storing morphosyntactic and semantic information. Each element of the FS is a feature-value pair (FVP). The value can be atomic or complex. A complex value is itself an FS. One of the features – the base feature, indicated in the call to MORPHE (the feature *stem*, by default) – must have a string value. For example, The feature structures (4) and (5) generate the English sentence *John eats an apple* and the Arabic nominative indefinite plural of the lexeme for the singular noun *mudarGisa0* 'teacher (fem.)' (plural: *mudarGisaAtM*), respectively:

(4)

```
((stem "eat")
(tense present)
(subject ((stem "John")
          (cat n)
          (agr ((number sg) (person 3)))))
(object ((stem "apple")
          (det (stem "a" (cat det)))
          (cat n)
(agr ((number sg) (person 3)))))
```

(5)

```
((stem "mudarGisaO") (cat n) (gender f) (number pl) (case nom)
(def -))
```

The choice of feature names and values is entirely up to the user of MORPHE. The FVPs in an FS come from one of two sources. Static features, such as *cat* (part of speech), *gender*, and *stem* come from the lexicon. Dynamic features such as *Case*, *def* (definiteness), and usually *number*, are determined by MORPHE's caller. In a machine translation context, the input sentence and linguistic knowledge in the target language would determine the exact contents of the FS passed to the morphological generator.

The output of MORPHE is simply a string with the desired transformations relative to the input *stem* value.

**<u>The Morphological Form Hierarchy:</u>**

MORPHE is based on the notion of a morphological form hierarchy (MFH) or tree that acts, in generation, as a discrimination network. Each internal node of the tree specifies a piece of the FS that is common to that entire subtree. The root of the tree is a special node that simply binds all subtrees together. The leaf nodes of the tree correspond to distinct morphological forms in the language. Each node in the tree below the root is added by specifying the parent of the node and the logical combination of FVPs that distinguish the node from its parent and siblings.

The syntax for declaring a morphological form is:

(morph-form <node name> <parent node name> <FVP combination>)

A possible (and incomplete) tree for English, showing nouns and verbs, is given in Figure 1 below. The notation for specifying the noun subtree is shown in (6) below:

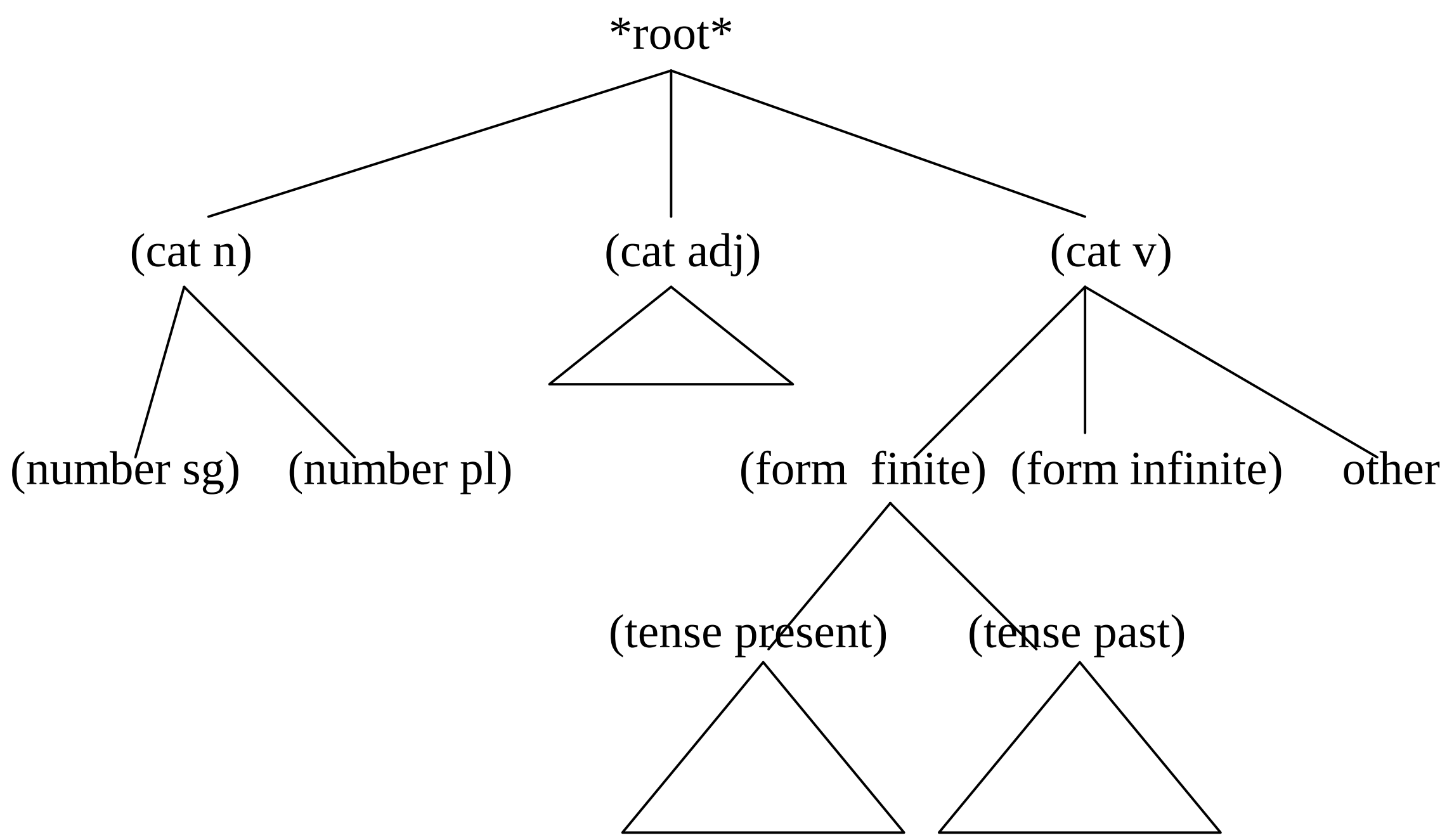

**Figure 1: A partial hierarchy for English Morphology**

(6)

```
 (morph-form n   * (cat n))
(morph-form n+sing n (number sg))
(morph-form n+plur (number pl))
```

For example, the declaration

```
 (morph-form n+sing n (number sg))
```

constructs the node *n+sing* as a child of *n*.

**Transformational Rules:**

A rule attached to each (leaf) node of the MFH effects the desired morphological transformations for that node. A rule consists of one or more mutually exclusive clauses. The *if* part of a clause is a regular expression pattern, which is matched against the string on which the rule must act. The *then* part includes one or more operators, applied in the given order. Operators include addition, deletion, and replacement of prefixes, infixes, and suffixes. The output of the transformation is the transformed string. The following rules illustrate MORPHE's Syntax.

**(i) Addition**

```
(morph-rule v-psfix-perf-3-sg-f
  (""
   (+s "at"))
  )
```

This rule suffixes the Arabic perfective third person, singular and feminine inflectional marker *at* to verbs. The operator used for addition is "+". "+s" (i.e., plus suffix) and "+p" (i.e., plus prefix) denote suffix addition and prefix addition, respectively. This rule has a null test, as is indicated by the absence of any context between the quotes "" (i.e., it applies everywhere).

**(ii) Replacement**

```
(morph-rule v-stem-f1-act-perf-1/2
("^%{cons}(a[wy]i)%{cons}$"
   (ri *1* "i")
  )
)
```

This rule is used to generate the active perfective first or second person form 1 stem for verbs whose patterns match the regular

Figure 2 below shows a portion of the computational implementation of noun generation in the extended version of MORPHE described above. The MFH is fully fleshed out only for nominative indefinite noun inflection, but the accusative and genitive cases and definite nouns are handled in a parallel way.[7] Also, while in the actual MFH all leaf and non-leaf nodes are given a name (ideally one that reflects the feature-value path used to reach them), the Figure shows node names only for the subset of nodes referenced in the following discussion; for the remainder of the nodes, only the distinguishing feature-value pairs are shown.

---

[7] Definite nouns (feature value pair (def +) ) have a different suffix and the prefix 'Al' which, depending on the initial consonant of the noun, may either have a *sukuwn* on the 'l' (i.e. 'Alo') or undergo assimilation and cause gemination of the initial consonant.

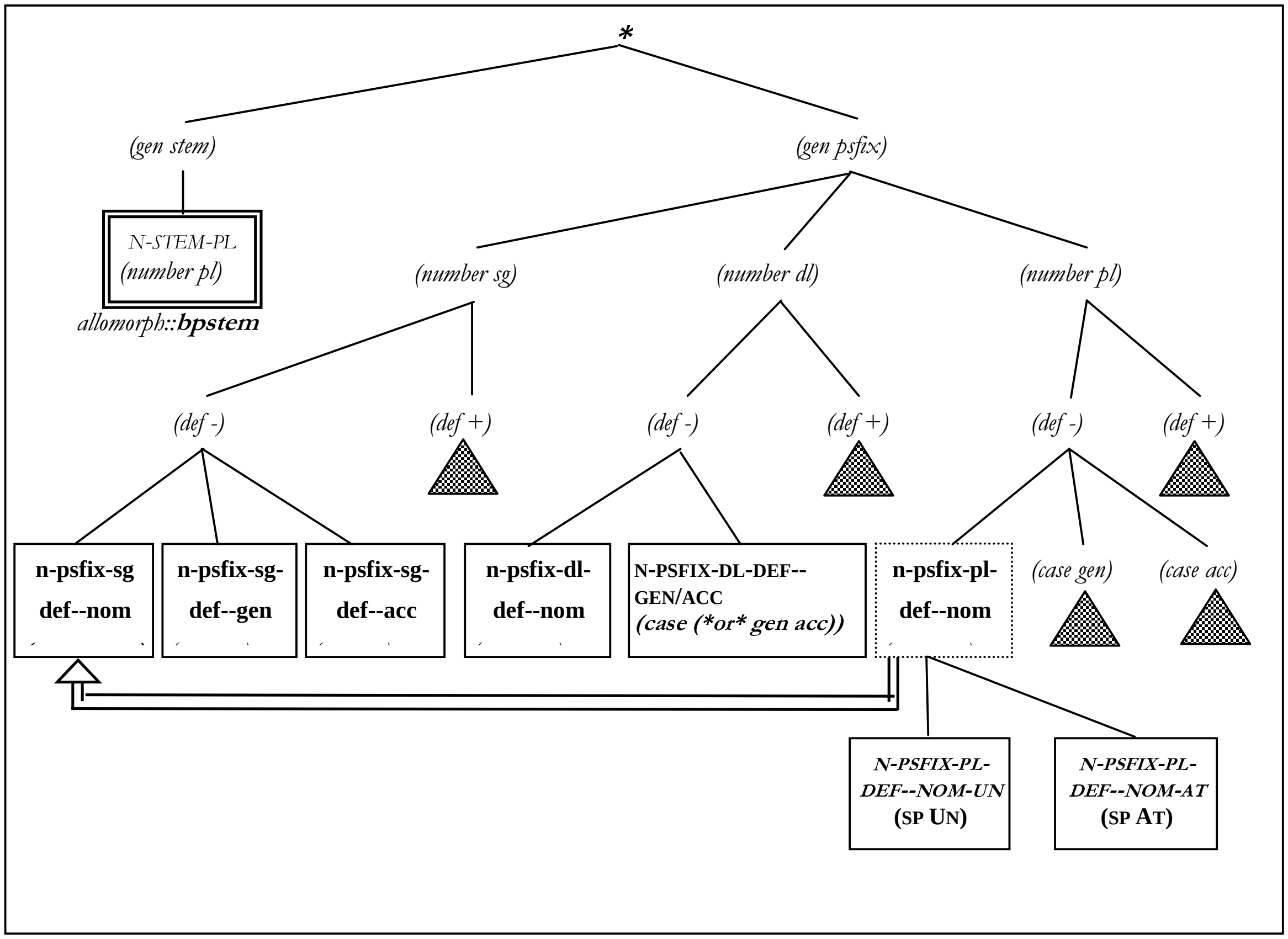

**Figure 2: The Noun Morphological Form Hierarchy (MFH)**

Generation of a fully inflected noun requires two calls to MORPHE, the first to generate the required stem, the second to generate the appropriate inflectional prefixes and suffixes. These two calls, effected by temporarily adding the FVPs *(gen stem)* and *(gen psfix)*, to the FS respectively, are reflected in the first level of branching below the root node *(*)* in the MFH. In the first call, *gen* is set to *stem*, so MORPHE traverses the *(gen stem)* subtree. For singular nouns, there is no stem change. Therefore, no branch for *(number sg)* appears in the MFH. MORPHE uses the feature *stem* as base feature. For plural nouns, an allomorph rule is attached to the node *N-STEM-PL* using the declaration:

(morph-allomorph n-stem-pl bpstem)

The effect of the allomorph rule is to consult the FS and return the value of the feature *bpstem*, if it is present, for subsequent operations. If there is no *bpstem* feature, MORPHE returns the value of the base feature stem. The value returned on this first call is used as the base feature value for the second call.

The second time MORPHE is called, the *gen* feature is set to *psfix* and MORPHE traverses the *(gen psfix)* branch of the MFH to add inflectional affixes to the stem obtained from the previous call. Consider first the simpler cases of singular and dual nouns. If MORPHE is given the FS for a singular indefinite noun, it will follow the leftmost branches of the subtree. Depending on the *Case* feature, it will reach one of the leaf nodes labeled *N-PSFIX-SG-DEF--NOM*, *N-PSFIX-SG-DEF--GEN*, *N-PSFIX-SG-DEF--ACC*. Each of these leaf nodes has attached to it a transformation rule that will add the appropriate suffix to the present stem. For example, the following rule adds the appropriate nominative indefinite plural suffix to nouns that form the sound plural in *aAt* (e.g., *mudarGisa0M* "teacher (fem)" (sg) → *mudarGisaAtM* (pl)).

```
(morph-rule n-psfix-pl-def--nom-aAt
  ("O$"
   (rs "O" "aAtM")
   )
  (""
   (+s "aAtM")
   ))
```

The syntax "O$" is used to match a 0 at the end of the input string, whereas the empty string "" matches anything. The "rs" (replace suffix) operator replaces its first argument with its second argument in the input string. The "+s" (add suffix) operator simply adds a suffix.

If MORPHE is given the FS for a dual indefinite noun, MORPHE will traverse the middle branch of the *(gen psfix)* subtree and add the appropriate dual ending. Note that genitive and accusative cases have the same suffixes in the dual, so they are treated as a single leaf node in the MFH.[8]

The case of plural nouns is more complex. Given a FS containing the feature-value pair *(case nom)* and *(def -)*, the appropriate suffixes depend on the plural class associated with the noun (see the rules governing the Arabic plural generation in section 2). The two classes of nouns with sound plurals are distinguished by the feature *sp* in the FS, with values *uwna* for the sound plural masculine (e.g., *mudarGis* (sg) → *mudarGisuwna* (pl) ) or *aAt* for the sound plural feminine (e.g., *mudarGisaO* (sg) → *mudarGisaAt*). If MORPHE finds the *sp* feature with one of these values in the FS, it will apply the transformational rule attached to the appropriate leaf *node* (*N-PSFIX-PL-DEF--NOM-UWNA* or N-PSFIX-PL-DEF--NOM-aAT).[9]

[8] The suffixes are *aAni* for nominative and *ayoni* for genitive and accusative for both indefinite and definite nouns. For nouns ending in *0*, the latter must be replaced with a regular *t* before the suffixes are added.

If the *sp* feature is missing, a broken plural is assumed, and MORPHE applies the default rule attached to the node *N-PSFIX-PL-DEF--NOM*. This rule is actually the same as the rule for nominative indefinite nouns, as indicated by the double arrow in Figure 2. The two rules are declared to be equivalent by the equivalence declaration:

```
(morph-equivalence n-psfix-sg-def--nom (n-psfix-pl-def--nom))
```

The actual transformational rule, shown above, is declared only once with the name *n-psfix-sg-def--nom*.

To clarify and illustrate the above description, consider the following two cases:

1. <u>Noun with a sound plural</u>

Input FS: *((stem "mudarGis") (sp uwna) (number pl) (case nom) (def -))*
In the first call to MORPHE, the *(gen stem)* subtree is traversed using the feature *(number pl)* and reaching the leaf node *N-STEM-PL*. MORPHE tries to apply the allomorph rule but, since no *bpstem* feature is found in the FS, the existing stem *mudarGis* is returned for use in the second call. When MORPHE is called again, the *(gen psfix)* subtree is traversed until the node labeled *N-PSFIX-PL-DEF--NOM-UWN* is reached. The attached transformational rule produces the result *AlomudarGisuwna* "the teachers":

```
(morph-rule n-psfix-pl-def+-nom-un
   (""
      (+p "Alo")
      (+s "uwna")
   )
)
```

The condition part of the rule is the empty string "", which matches any input. The operator "+p" prefixes the definite article *Alo* and the operator "+s" suffixes *uwna*. Note here that we are using the number exponent and the Case exponent as a portmanteau morpheme.

2. <u>Noun with a broken plural</u>

Input FS: *((stem "rajul") (bpstem "rijaAl") (number pl) (case nom) (def -))*

[9] The rules add suffixes *uwna* (nominative) or *iyna* (genitive and accusative cases) for nouns with the feature-value pair (sp uwna), suffix *aAtM* (nominative) or *aAtK* (genitive and accusative cases) for nouns with the feature-value pair (sp aAt). For nouns ending in *0*, the latter is removed before the suffixes are added.

In the first call to MORPHE, the *(gen stem)* subtree is traversed, reaching the leaf node *N-STEM-PL*. The value of the feature *bpstem – rijaAl* "men"– is retrieved from the FS and returned for use in the second call. Note that the broken plural stem is retrieved by the allomorph rule that is attached to the node *N-STEM-PL*. When MORPHE traverses the second part of the hierarchy, it reaches the node *N-PSFIX-PL-DEF--NOM*. No *sp* feature is found in the feature structure, so MORPHE defaults to the rule attached to *N-PSFIX-PL-DEF—NOM*, which is the same rule attached to node *N-PSFIX-SG-DEF--NOM* shown above.

Two related observations regarding the design of the MFH with respect to broken plurals are in order. First, while it would have been more convenient to create the distinction between the two sound plurals and the broken plurals further up in the *(gen psfix)* subtree, immediately under the *(number plural)* node, and equivalence the entire broken plural subtree of the *(number singular)* subtree, the current implementation of the enhanced MORPHE system requires that only leaf and pre-leaf nodes be equivalenced.[10] This limitation makes it necessary to push the default rules for broken plurals to the bottom of the tree. Second, since the genitive and accusative of sound plurals have the same surface realization, one may wonder why separate branches are shown for the genitive and the accusative in the (number plural) subtree. The three-way branching for Cases is made necessary by the broken plural default rules: broken plurals like singulars, do not have the same realization in the genitive and accusative and therefore separate equivalence rules must be created for each case. For sound plurals, the nodes/rules for the accusative are, in fact, equivalenced to the corresponding genitive node for both values of the definiteness feature.

In this paper, we have shown that a multiple-stem approach to the Arabic highly allomorphic broken plural system, couched within the lexeme-based morphology, obviates the need to posit complex and multiple rules in the generation of Arabic nouns. We have also provided an implementation of the linguistic analysis in Morphe. In this context, we have shown how the generalizations discussed in the linguistic analysis are captured in Morphe using equivalencing rules.

## **Transliteration**

The transliteration we use is for the sake of the sake of implementation and portability. This is not a phonetic system.

| | |
|---|---|
| ا(?alif) | A |
| ب | b |
| ت | t |
| ث | # |

[10] More accurately, since it is rules that are actually equivalenced, only those nodes that may have rules attached to them, i.e. leaf and pre-lead nodes, can be equivalenced.

| | |
|---|---|
| ج | j |
| ح | H |
| خ | x |
| د | d |
| ذ | › |
| ر | r |
| ز | z |
| س | s |
| ش | š |
| ص | S |
| ض | D |
| ط | T |
| ظ | Z |
| ع | ` |
| غ | < |
| ف | f |
| ق | q |
| ك | k |
| ل | l |
| م | m |
| ن | n |
| ه | h |
| و | w |
| ي | y |
| ى | Y |
| ة | 0 |
| َ(fatHa) | a |
| ِ(kasra) | i |
| ً(fatHtaAn) | F |
| ٍ(kasrataAn) | K |
| ’(DamGa) | u |
| ’’(DamGataAn) | M |
| ◦(sukuwun) | ° |
| ّ(šadGa) | G |
| أ(hamza on ^alif) | ^ |
| إ(hamza under ^alif) | ç |
| ٱ(waSla0 on ^alif) | ~ |
| ؤ(hamza on waaw) | V |
| ﺋ(hamza on line) | @ |
| ئ(hamza on yaA?) | v |